\pdfoutput=1
\documentclass{article}
\usepackage[T1]{fontenc}
\usepackage{iclr2027_conference,times}

\usepackage{amsmath,amsfonts,bm}

\def\eqref#1{equation~\ref{#1}}

\def\1{\bm{1}}

\def\va{{\bm{a}}}
\def\vb{{\bm{b}}}

\def\vg{{\bm{g}}}
\def\vh{{\bm{h}}}

\def\vm{{\bm{m}}}

\def\vp{{\bm{p}}}

\def\vr{{\bm{r}}}
\def\vs{{\bm{s}}}

\def\vx{{\bm{x}}}
\def\vy{{\bm{y}}}
\def\vz{{\bm{z}}}

\DeclareMathAlphabet{\mathsfit}{\encodingdefault}{\sfdefault}{m}{sl}
\SetMathAlphabet{\mathsfit}{bold}{\encodingdefault}{\sfdefault}{bx}{n}

\usepackage{amsmath,amssymb,booktabs,graphicx,microtype,multirow,url,xcolor,hyperref,enumitem,flafter,placeins}
\definecolor{acrossblue}{RGB}{30,90,170}
\definecolor{acrossteal}{RGB}{17,139,139}

\newcommand{\valpha}{\boldsymbol{\alpha}}
\hypersetup{colorlinks=true,citecolor=acrossblue,linkcolor=acrossblue,urlcolor=acrossteal}
\setlist[itemize]{leftmargin=*,topsep=2pt,itemsep=1pt}

\title{AcrossVAM1.0: Particle World Modeling\\ for Text-Assisted Robot Video Prediction}
\author{Yafei Zhang$^{\,1,2}$ \quad Nan Wu$^{\,1}$\thanks{Corresponding author.}\\
\normalfont $^{1}$ Across Physical AI, Beijing, China\\
\normalfont $^{2}$ Institute of Automation, Chinese Academy of Sciences, Beijing, China\\
\normalfont \texttt{18453881970@163.com}, \texttt{across2026@163.com}}
\iclrfinalcopy

\begin{document}
\raggedbottom
\maketitle
\lhead{Preprint.}

\begin{abstract}
Predicting robot videos requires both precise motion reasoning and preservation of
high-frequency appearance, yet monolithic pixel models entangle these objectives
and often conceal their progress behind a strong last-frame baseline.
We present \textbf{AcrossVAM1.0}, a lightweight, text-assisted video action model that
factorizes future prediction into object-centric motion and dense appearance.
A frozen SAM3-DLP codec decomposes four context frames into semantic particles for
the robot, arm, and gripper, together with a background latent. A 0.28M-parameter
spatio-temporal Transformer aligns particle identities, rolls their states forward,
and is modulated by a frozen OpenCLIP instruction embedding through FiLM.
A causal dual-stream decoder combines particle-rendered motion with appearance
encoded exclusively from the last observed frame; a residual refiner and learned
delivery mask produce five future frames without access to future appearance.
On our VRS benchmark constructed from diverse real-robot trajectories, particle
dynamics reduce trajectory error by 21.0\% over persistence. Across three delivery-mask
seeds, AcrossVAM1.0 improves future-frame PSNR/SSIM from 19.97/0.796 to
$20.573{\pm}0.009$/$0.8004{\pm}0.0002$, while raw particle generation improves
motion-region PSNR from 11.89 to 13.23. The delivered model does not yet beat
persistence in LPIPS ($0.1304{\pm}0.0004$ versus 0.122), and correct-versus-
shuffled language changes trajectory error by only 2.8--3.1\%. We report these
limitations alongside oracle, negative-control, multi-seed, and per-robot analyses.
The results show that explicit particle dynamics are a promising low-dimensional
interface for robot video prediction, while robust language grounding and
appearance delivery remain the principal open challenges.
\end{abstract}

\section{Introduction}
\label{sec:intro}

A predictive model of how a scene will evolve is useful for robot planning,
representation learning, and data generation. Video prediction offers a general
interface because it retains the geometry and appearance needed by downstream
visual policies~\citep{finn2016unsupervised,ebert2018visual,du2023unipi}. However,
real robot videos present an uncomfortable combination: most pixels are static,
small articulated parts determine the action, and the same natural-language
instruction can admit visually distinct trajectories. A model can therefore achieve
competitive global image metrics by copying the last frame, without learning the
motion that matters.

Contemporary world models typically compress the whole frame into a dense latent
and learn its temporal evolution~\citep{hafner2023dreamerv3,babaeizadeh2021fitvid,
huang2026nano}. This representation is effective but leaves correspondence,
dynamics, and texture synthesis coupled. Object-centric models instead represent a
scene as slots or particles~\citep{locatello2020slot,wu2023slotformer,daniel2023ddlp,daniel2026latent}.
Their explicit state makes motion easier to inspect, but a compact object bottleneck
cannot reconstruct fine texture by itself. Moreover, conditioning a video model on
language does not guarantee that its trajectory depends on the instruction.

\begin{figure*}[t]
  \centering
  \includegraphics[width=\textwidth]{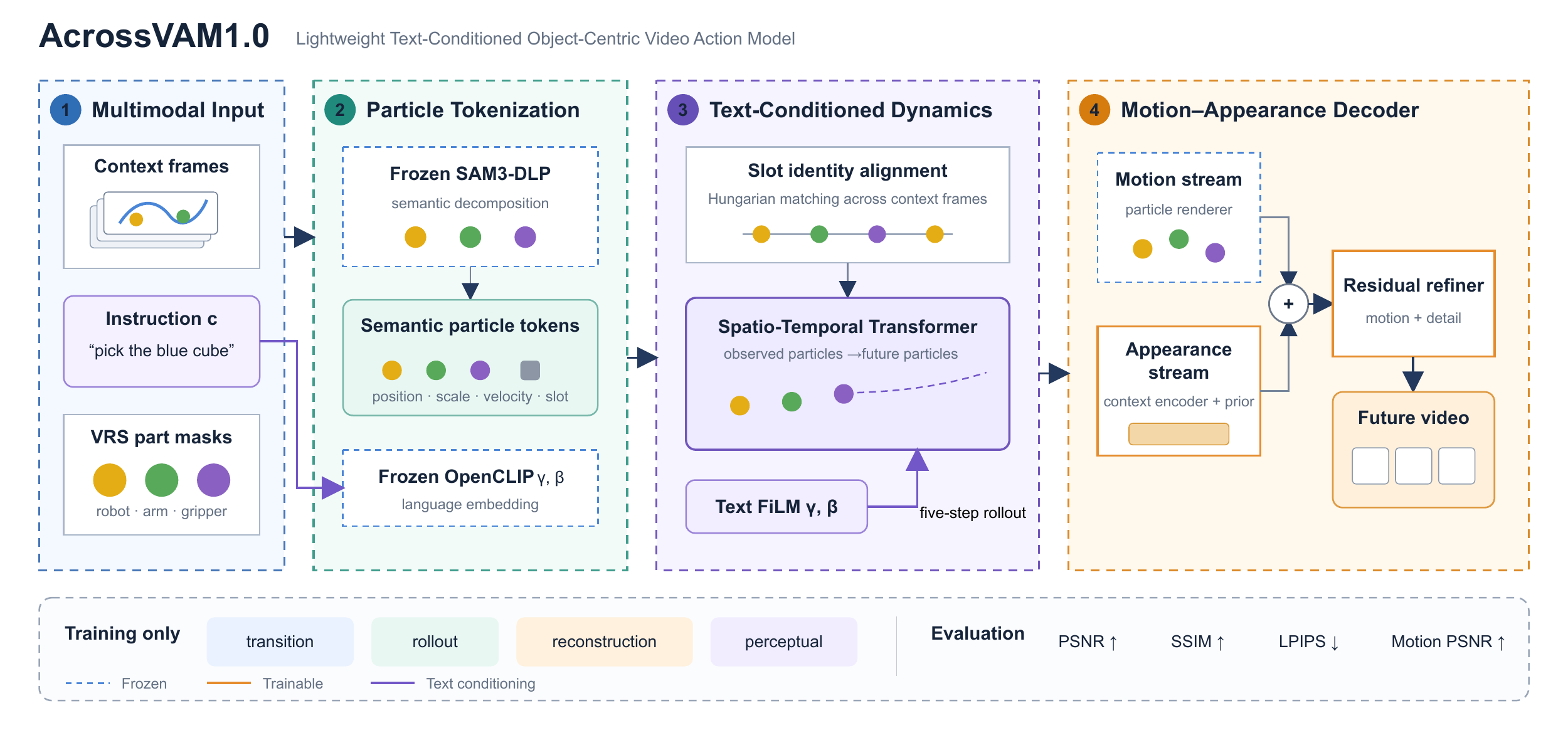}
  \caption{\textbf{AcrossVAM1.0 overview.} Four context frames and an instruction are
  encoded into semantic robot particles and a background latent. Identity-aligned,
  text-modulated particle dynamics predict five future states. A frozen renderer
  produces motion structure, while a context-only dense stream supplies appearance;
  the residual delivery module fuses both predictions. Snowflakes and flames denote
  frozen and trainable modules. Dashed gray paths are training-only.}
  \label{fig:overview}
\end{figure*}

AcrossVAM1.0 assigns different responsibilities to different representations
(Figure~\ref{fig:overview}). Semantic particles describe the geometry of the whole
robot, arm, and gripper; a compact Transformer predicts their future states; and a
separate causal appearance stream restores detail. Language is injected only into
the dynamics through feature-wise modulation, which makes its effect measurable at
the trajectory level. The dense branch observes the last context frame but never a
future frame, so future motion cannot leak through the appearance path.

Our contributions are:
\begin{itemize}
  \item We introduce a semantically anchored particle world model that reduces
  real-robot video dynamics to a compact trajectory problem. Its dynamics core has
  only 0.28M trainable parameters and exposes part-level motion for diagnosis.
  \item We propose a causal motion--appearance decoder that combines predicted
  particle geometry with context-only dense appearance and explicitly controls the
  trade-off between sharp static regions and moving-object fidelity.
  \item We evaluate against persistence, shuffled-language and random controls,
  oracle particles, multiple seeds, and per-robot splits. The evaluation identifies
  both a reliable motion benefit and the still-limited degree of language grounding.
\end{itemize}

\section{Related Work}
\label{sec:related}

\paragraph{Video prediction for physical interaction.}
Video prediction has long served as a test bed for learning physical regularities
without manual state annotation. Early action-conditioned models predicted future
pixels from images and robot controls, demonstrating that visual foresight can
support planning and representation learning~\citep{finn2016unsupervised,ebert2018visual}.
This line of work is especially relevant to manipulation because the prediction
target preserves object geometry, contact events, and scene appearance in a common
observation space. It also exposes a characteristic evaluation problem: robot
videos contain large static regions, so copying the last context frame can score
well even when articulated motion is wrong. Recent text-guided approaches instead
use language to describe a desired evolution and treat predicted video as a
policy-like interface~\citep{du2023unipi}. AcrossVAM1.0 follows this conditional
formulation but explicitly reports both global image quality and motion-region
quality, making improvement over persistence harder to obtain by static copying.

\paragraph{Latent and diffusion world models.}
Dense video predictors have progressively combined stochastic latent variables,
recurrent state, and learned feature spaces to model uncertainty while reducing the
cost of direct pixel prediction. FitVid shows that a carefully regularized latent
video model remains a strong reference for stochastic prediction
~\citep{babaeizadeh2021fitvid}. Learned latent world models can also optimize
control-relevant dynamics without reconstructing every observation at every step
~\citep{hafner2023dreamerv3}. Diffusion models provide a complementary route to
multimodal generation~\citep{ho2020ddpm}; Diffusion Forcing extends this view by
training sequence models with independently corrupted tokens and flexible
conditioning patterns~\citep{chen2024diffusionforcing}. These methods obtain
expressive dense representations, but motion, correspondence, and high-frequency
appearance typically remain entangled in the same latent process. Nano World Models
ask how much predictive structure can be retained in deliberately small models
~\citep{huang2026nano}. AcrossVAM1.0 shares this efficiency objective while placing
a semantic particle bottleneck between perception and rendering: a small dynamics
core predicts geometry, whereas a separate causal stream supplies texture from
observed context.

\paragraph{Object-centric representation and dynamics.}
Object-centric learning decomposes a scene into reusable entities rather than one
global feature map. Slot Attention learns exchangeable component representations
from images~\citep{locatello2020slot}, and SAVi++ extends slot-based perception to
video with temporal propagation and stronger supervision signals
~\citep{elsayed2022saviplus}. SlotFormer subsequently predicts future slot states
with a Transformer, showing that temporal reasoning can be separated from the
decoder~\citep{wu2023slotformer}. Particle representations make the spatial state
more explicit: DDLP associates appearance with position, scale, depth, and presence,
while later latent-particle models study long-horizon dynamics in this structured
space~\citep{daniel2023ddlp,daniel2026latent}. The benefit is interpretability and a
compact motion interface; the cost is that exchangeable entities require temporal
correspondence and may discard texture needed for photorealistic rendering. Set
prediction commonly addresses permutation ambiguity with bipartite assignment
~\citep{carion2020detr}. AcrossVAM1.0 specializes these ideas to articulated robots:
VRS supervision anchors whole-robot, arm, and gripper particles, Hungarian matching
supports unordered proposals, and a dense residual code restores appearance that
should not be forced through the low-dimensional motion state.
Recent extensions sharpen both axes: OCK explicitly augments object slots with
position, velocity, and acceleration~\citep{song2025ock}, while TextOCVP uses a
text-conditioned slot Transformer for language-guided prediction~\citep{villar2026textocvp}.

\paragraph{Foundation perception and language conditioning.}
Large frozen encoders can provide semantic structure without requiring an equally
large trainable predictor. Promptable segmentation models such as SAM and SAM2
transfer object masks across diverse scenes and video~\citep{kirillov2023sam,ravi2024sam2}.
CLIP and OpenCLIP provide transferable text representations aligned with visual
concepts~\citep{radford2021clip,cherti2023openclip}, while FiLM injects a conditioning
signal through feature-wise affine modulation~\citep{perez2018film}. These
components make language conditioning inexpensive, but their presence alone does
not demonstrate that a predicted trajectory depends on the instruction: a model
may rely on scene priors, dataset bias, or the observed motion prefix. We therefore
freeze perception and language towers, apply FiLM only to particle dynamics, and
evaluate correct text against shuffled, pseudo-text, and random-label controls.
This design separates semantic perception from causal language use and motivates
our conservative term \emph{text-assisted} rather than instruction-controlled.

\paragraph{Positioning of AcrossVAM1.0.}
Prior work generally emphasizes either dense visual fidelity, compact latent
dynamics, or object-centric interpretability. AcrossVAM1.0 combines these aims by
assigning them to distinct modules: frozen semantic perception defines particles,
a 0.28M-parameter Transformer models their motion, and a context-only appearance
path performs residual delivery. The closest comparisons are therefore
complementary rather than interchangeable: dense predictors test image quality,
slot and particle models test structured dynamics, and language-conditioned video
models test semantic responsiveness. Our evaluation mirrors this decomposition
with persistence, motion-region, particle-oracle, language-control, multi-seed, and
cross-robot diagnostics.
\section{Method}
\label{sec:method}

\subsection{Problem formulation}
Let $\vx_{1:C}=(\vx_1,\ldots,\vx_C)$ be context RGB frames and $c$ a natural-language
instruction. We predict $\hat{\vx}_{C+1:T}$, with $C=4$ and $T=9$. AcrossVAM1.0
factorizes the conditional distribution as
\begin{equation}
 p(\vx_{C+1:T}\mid \vx_{1:C},c)
 = \int p_{\phi}(\vx_{C+1:T}\mid \vz_{C+1:T},\vx_C)
 p_{\theta}(\vz_{C+1:T}\mid \vz_{1:C},c)\,d\vz ,
\label{eq:factor}
\end{equation}
where $\vz$ is an object-centric state. Dynamics $p_\theta$ is responsible for
motion; decoder $p_\phi$ renders motion and context appearance.

\subsection{Frozen semantic particle codec}
For each frame, a frozen SAM3-DLP encoder receives RGB and VRS part masks and
extracts $K=3$ semantic particles, corresponding to the whole robot, arm, and
gripper, plus a spatial background latent $\vb_t$. Particle $i$ is
\begin{equation}
 \vz_t^i=[\vp_t^i,\vs_t^i,d_t^i,q_t^i,\va_t^i],
\end{equation}
where $\vp\in[0,1]^2$ is position, $\vs\in\mathbb{R}_+^2$ is scale, $d$ is depth,
$q$ is presence, and $\va$ is appearance. Explicit geometry forms the prediction
target; appearance and background serve the renderer. Freezing the codec prevents
the representation from drifting toward an easy pixel-copying solution.

\paragraph{Identity alignment.}
Unordered proposals require temporal correspondence. For adjacent frames, we form
\begin{equation}
 C_{ij}=\lambda_p\|\bar{\vp}_{t-1}^i-\bar{\vp}_t^j\|_2
 +\lambda_a\left(1-\cos(\va_{t-1}^i,\va_t^j)\right)
\end{equation}
and obtain $\pi_t=\arg\min_{\pi}\sum_i C_{i,\pi(i)}$ using Hungarian matching.
The bar denotes scale-normalized coordinates. In our offline VRS cache, fixed
semantic part identifiers already determine the permutation; we bypass matching
there and retain it for proposal-based inference.

\subsection{Text-assisted particle dynamics}
The dynamics input concatenates geometry, first-order velocities
$\Delta\vp_t^i,\Delta\vs_t^i$, depth, presence, and learned slot and time embeddings.
A causal spatio-temporal Transformer~\citep{vaswani2017attention} alternates temporal
reasoning within each slot with interactions among robot parts. It predicts residual
updates autoregressively:
\begin{equation}
 \hat{\vy}_{t+1}^i=\hat{\vy}_t^i+
 f_{\theta}(\hat{\vy}_{1:t}^{1:K},e_c)^i,\qquad
 \vy=[\vp,\vs,d,q].
\label{eq:dynamics}
\end{equation}
Here $e_c$ is produced by a frozen OpenCLIP ConvNeXt-L/320 text tower. At block
$\ell$, trainable FiLM maps inject instruction semantics:
\begin{equation}
 \tilde{\vh}_{\ell}=
 (1+\gamma_{\ell}(e_c))\odot\operatorname{LN}(\vh_\ell)
 +\beta_{\ell}(e_c).
\label{eq:film}
\end{equation}
FiLM projections are initialized to zero, so optimization starts from unconditioned
dynamics rather than a randomly perturbed model. We combine one-step transition and
autoregressive rollout supervision:
\begin{equation}
 \mathcal{L}_{\mathrm{dyn}}=
 \lambda_{\mathrm{tr}}\sum_{t=C}^{T-1}\|\hat{\vy}_{t+1}^{\mathrm{TF}}-\vy_{t+1}\|_1+
 \lambda_{\mathrm{ro}}\sum_{t=C+1}^{T}\|\hat{\vy}_{t}^{\mathrm{AR}}-\vy_t\|_1.
\end{equation}

\subsection{Causal motion--appearance decoding}
The frozen particle renderer maps predicted particles and context background to a
structural RGB proposal $\vr_t$ and alpha map $\valpha_t$. This proposal moves the
correct parts but lacks high-frequency detail. In parallel, a dense encoder extracts
$\vh_C=E_{\mathrm{app}}(\vx_C)$ from the \emph{last context frame only}. A conditional
residual prior predicts future detail from context appearance and particle motion:
\begin{equation}
 \hat{\vh}_t=P_{\psi}(\vh_C,\vr_C,\valpha_C,\vr_t,\valpha_t,t-C).
\end{equation}
A residual refiner synthesizes
$\vg_t=R_{\phi}(\vr_t,\valpha_t,\hat{\vh}_t,\vx_C)$. Finally, a learned confidence
mask $\vm_t\in[0,1]^{H\times W}$ blends generation with persistence:
\begin{equation}
 \hat{\vx}_t=\vm_t\odot\vg_t+(1-\vm_t)\odot\vx_C .
\label{eq:blend}
\end{equation}
The mask uses inference-time quantities such as predicted alpha motion and candidate
disagreement. The appearance loss is
\begin{equation}
 \mathcal{L}_{\mathrm{app}}=\lambda_1\|\vg_t-\vx_t\|_1+
 \lambda_s(1-\operatorname{SSIM}(\vg_t,\vx_t))+
 \lambda_p\operatorname{LPIPS}(\vg_t,\vx_t),
\end{equation}
with extra motion-region weighting during training. Future RGB, masks, and residual
codes are targets only. At inference every decoder input is a function of
$(\vx_{1:C},c,\hat{\vz}_{C+1:T})$; the dense branch cannot leak future appearance.

\section{Experiments}
\label{sec:experiments}

\subsection{Setup}
\paragraph{Data and protocol.}
We construct the Video Robot Segmentation (VRS) benchmark from diverse DROID-style real-robot trajectories
~\citep{khazatsky2024droid}, augmented with masks for whole robot, arm, and gripper
and a natural-language command. The manifest contains 2,746 labeled clips:
2,526/106/105 are assigned to train/validation/test and nine records are excluded
from the fixed protocol. We resize frames to $128\times128$, sample nine-frame
windows at stride three, observe four frames, and predict five. The future-video
test contains 92 clips and 460 predicted frames; codec reconstruction uses 48 clips
and 432 frames; trajectory diagnostics use 368 held-out windows. Platforms include
Franka, WidowX, Mobile ALOHA, Google Robot, Fanuc, Sawyer, Stretch, UR5, Kuka IIWA,
and xArm.

\paragraph{Metrics and selection.}
We report PSNR, SSIM~\citep{wang2004ssim}, and LPIPS~\citep{zhang2018lpips} over all
future pixels, plus PSNR in a dilated motion/foreground region. Checkpoints and
blending thresholds are selected on validation data; test is evaluated once.
Persistence copies $\vx_C$. The particle oracle uses ground-truth future particles,
and the pixel oracle chooses generation or persistence per pixel; both are upper
bounds, not deployable methods.

\paragraph{Implementation.}
The particle Transformer has width 64, two layers, four heads, and 0.28M parameters.
The complete trainable future-prediction stack has approximately 9.5M parameters,
excluding frozen SAM3-DLP and OpenCLIP backbones. We train codec, dynamics,
appearance, and delivery modules in stages. Appendix~\ref{app:details} gives details.

\subsection{Future video prediction}
\begin{table}[t]
\caption{Future prediction on 92 held-out clips (460 future frames). AcrossVAM1.0
reports mean $\pm$ sample standard deviation over three independently initialized
delivery masks; upstream dynamics, appearance, and diffusion-prior checkpoints are
shared. Only results evaluated on the identical VRS protocol are included; scores
reported on other datasets are not transplanted. Bold marks the best evaluated
deployable result.}
\label{tab:video}
\centering\scriptsize
\begin{tabular}{lcccc}
\toprule
Method & PSNR$\uparrow$ & SSIM$\uparrow$ & LPIPS$\downarrow$ & Motion PSNR$\uparrow$\\
\midrule
Persistence & 19.97 & 0.796 & \textbf{0.122} & 11.89\\
Particle-conditioned generation & 19.93 & 0.763 & 0.138 & \textbf{13.23}\\
\quad $+$ fixed particle-alpha delivery & 20.17 & 0.785 & 0.134 & 12.93\\
\textbf{AcrossVAM1.0 DiffPrior (3 seeds)} & $\mathbf{20.573{\pm}0.009}$ &
$\mathbf{0.8004{\pm}0.0002}$ & $0.1304{\pm}0.0004$ & $12.870{\pm}0.032$\\
\midrule
Per-pixel selection oracle & 22.12 & 0.833 & 0.114 & 14.09\\
\bottomrule
\end{tabular}
\end{table}

Table~\ref{tab:video} exposes the central trade-off. Raw particle-conditioned
generation improves motion PSNR by 1.34 dB over persistence, but degrades global
SSIM because it modifies easy static pixels. Residual delivery recovers static
fidelity and improves global PSNR by 0.61 dB and SSIM by 0.0044 while retaining
most of the motion advantage. Dispersion across delivery-mask seeds is small
(0.009 dB PSNR and 0.0002 SSIM), but this sweep shares all upstream checkpoints and
therefore measures delivery robustness rather than full end-to-end uncertainty.
LPIPS remains 0.0082 worse than persistence. The 1.55 dB oracle gap indicates that
candidate quality is not the only bottleneck: deciding where to synthesize remains
imperfect.

\paragraph{Recent public baselines.}
We surveyed public methods released in the past five years: the dense stochastic
predictor FitVid, slot-based SlotFormer, particle-based DDLP, slot-based TextOCVP,
and particle-based LPWM. All provide public implementations, but none reports
the fixed VRS split, four-frame context, five-frame horizon, and motion-region metric
used here. We therefore separate measured results (Table~\ref{tab:video}) from the
reproduction queue (Table~\ref{tab:baseline-protocol}) rather than mixing
incommensurate numbers from BAIR, CLEVRER, OBJ3D, or LanguageTable. Reproduction
follows the priority LPWM-Language, TextOCVP, FitVid, DDLP, and SlotFormer.

\begin{table}[t]
\caption{Public-baseline reproduction protocol. This table tracks comparability
and execution status, not quantitative results. Every method will be retrained on
the fixed VRS split with four context and five future frames and evaluated on the
same 92 clips and 460 future frames as Table~\ref{tab:video}.}
\label{tab:baseline-protocol}
\centering\scriptsize
\begin{tabular}{cllll}
\toprule
Priority & Method & Published & Conditioning & Structured state / status\\
\midrule
1 & LPWM-Language~\citep{daniel2026latent} & 2026 & text & particles / partial: epochs 0--3 saved\\
2 & TextOCVP~\citep{villar2026textocvp} & 2026 & text & slots / queued\\
3 & FitVid~\citep{babaeizadeh2021fitvid} & 2022 & frames & dense latent / queued\\
4 & DDLP~\citep{daniel2023ddlp} & 2024 & frames & particles / queued\\
5 & SlotFormer~\citep{wu2023slotformer} & 2023 & frames & slots / queued\\
\bottomrule
\end{tabular}
\end{table}
\subsection{Qualitative behavior and delivery robustness}
\label{sec:qualitative}

\begin{figure}[t]
  \centering
  \begin{tabular}{@{}cc@{}}
    \includegraphics[width=0.485\linewidth]{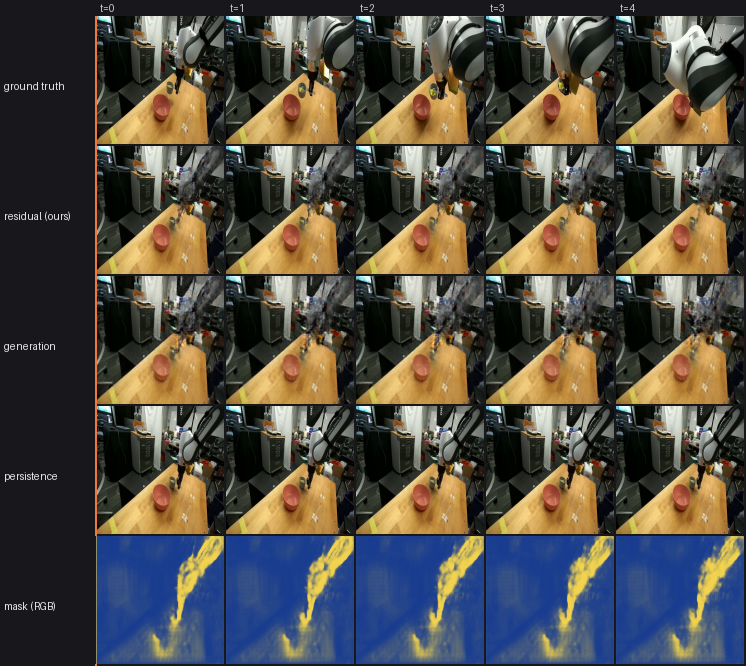} &
    \includegraphics[width=0.485\linewidth]{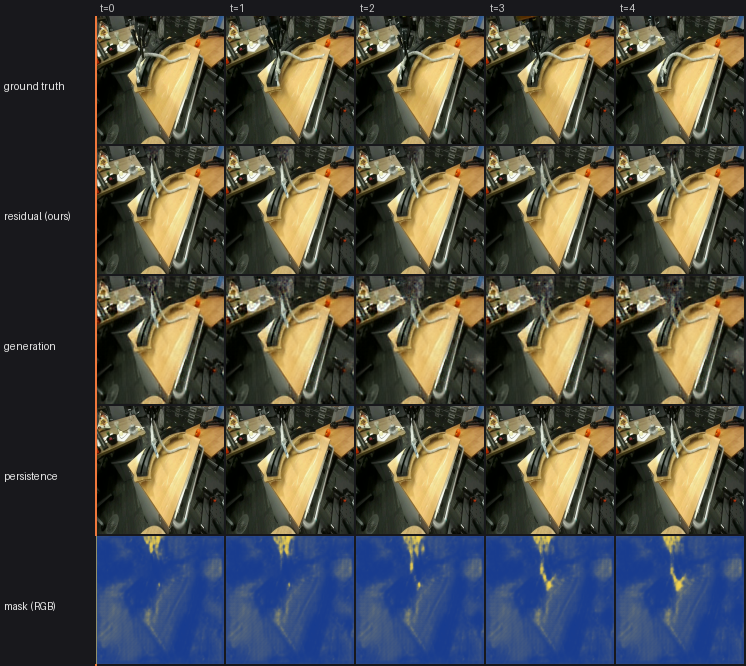}
  \end{tabular}
  \caption{\textbf{Representative held-out Franka predictions.} Each panel shows
  five future steps; rows are ground truth, residual delivery, raw generation,
  persistence, and the learned blend mask visualized with a blue-to-yellow RGB colormap. The left example has a compact interaction
  region, whereas the right example contains heavier clutter and occlusion. These
  cases are selected from the fixed seed-0 demo set after checkpoint selection;
  they are not used for quantitative model selection.}
  \label{fig:qualitative}
\end{figure}

The examples in Figure~\ref{fig:qualitative} illustrate why global and motion
metrics must be read together. Raw generation changes the articulated region but
also perturbs static texture. The blend mask localizes many edits, yet errors remain
near thin links, occlusion boundaries, and newly revealed pixels. The qualitative
failure pattern is consistent with the gap to the pixel-selection oracle.

\begin{table}[t]
\caption{Earlier DensePrior delivery-stage robustness over three appearance/refinement seeds on the
same 460-frame test set. The shared particle dynamics checkpoint is frozen, so this
is not a full end-to-end seed sweep. Values are mean $\pm$ sample standard deviation.}
\label{tab:delivery-seeds}
\centering
\scriptsize
\begin{tabular}{lcccc}
\toprule
Method & PSNR$\uparrow$ & SSIM$\uparrow$ & LPIPS$\downarrow$ & Motion PSNR$\uparrow$\\
\midrule
Persistence & 19.966 & 0.796 & \textbf{0.122} & 11.894\\
Raw DensePrior generation & $19.540\pm0.274$ & $0.718\pm0.018$ &
$0.160\pm0.011$ & $\textbf{13.057}\pm0.077$\\
AcrossVAM1.0 joint residual & $\textbf{20.464}\pm0.112$ &
$0.795\pm0.005$ & $0.129\pm0.001$ & $12.911\pm0.082$\\
\midrule
Pixel-selection oracle & $21.940\pm0.071$ & $0.830\pm0.001$ &
$0.112\pm0.001$ & $13.907\pm0.074$\\
\bottomrule
\end{tabular}
\end{table}

The earlier DensePrior variant retains a $0.50$ dB average PSNR gain over persistence and a
$1.02$ dB motion-region gain, with low dispersion. SSIM is statistically near the
persistence baseline and LPIPS remains worse. One seed selected its initialization
under the validation rule, further indicating that appearance refinement is less
stable than particle dynamics.

\begin{table}[t]
\caption{Exact DiffPrior delivery sweep. Each learned mask is independently
initialized and selected by validation SSIM; the test set is evaluated once after
selection. Particle rollouts, appearance features, and DiffPrior candidates are
shared across seeds.}
\label{tab:diffprior-seeds}
\centering
\scriptsize
\begin{tabular}{lccccc}
\toprule
Seed & Selected step & PSNR$\uparrow$ & SSIM$\uparrow$ & LPIPS$\downarrow$ & Motion PSNR$\uparrow$\\
\midrule
0 & 1,500 & 20.563 & 0.8006 & 0.1305 & 12.833\\
1 & 3,000 & 20.578 & 0.8004 & 0.1299 & 12.886\\
2 & 1,500 & 20.578 & 0.8002 & 0.1307 & 12.890\\
\midrule
Mean $\pm$ s.d. & -- & $20.573{\pm}0.009$ & $0.8004{\pm}0.0002$ &
$0.1304{\pm}0.0004$ & $12.870{\pm}0.032$\\
\bottomrule
\end{tabular}
\end{table}

The stronger DiffPrior delivery is similarly stable across mask initialization:
all three seeds improve over persistence in PSNR and SSIM, and the selected
checkpoints occur at 1,500--3,000 steps (Table~\ref{tab:diffprior-seeds}). Its
motion-region PSNR remains $0.98$ dB above persistence but $0.36$ dB below raw
generation, quantifying the cost of suppressing changes outside the predicted
delivery region. Because the upstream particle and appearance models remain fixed,
we report this result as a delivery-stage sweep rather than implying full
end-to-end multi-seed evidence.

\subsection{Representation and dynamics}
\begin{table}[t]
\caption{Representation and dynamics diagnostics. Float counts measure the extra
dense code per frame; improvement is trajectory-error reduction over persistence.}
\label{tab:representation}
\centering\small
\begin{tabular}{lrrrr}
\toprule
Representation & Floats & PSNR$\uparrow$ & SSIM$\uparrow$ & LPIPS$\downarrow$\\
\midrule
Base particle codec & 160 & 20.94 & 0.637 & 0.362\\
Generic dense code & 1,024 & 23.62 & 0.747 & 0.310\\
Particle + residual code & 4,096 & \textbf{27.95} & \textbf{0.877} & \textbf{0.148}\\
\bottomrule
\end{tabular}
\vspace{3pt}

\begin{tabular}{lcc}
\toprule
Dynamics model & Val improvement & Test improvement\\
\midrule
Width-64, 2-layer particle Transformer & 14.1\% & \textbf{21.0\%}\\
\quad three-seed test range & -- & [20.6\%, 21.4\%]\\
\bottomrule
\end{tabular}
\end{table}

Adding a $32\times32\times4$ residual code raises reconstruction PSNR by 7.01 dB
(Table~\ref{tab:representation}), motivating a dual stream rather than forcing
texture through object state. The small dynamics model reliably beats trajectory
persistence across seeds. Diagnostics show that residualized box coordinates and
raw appearance features introduce shortcuts; removing them and using stride-three
motion raises error reduction from low single digits to 21.0\%.

\subsection{Does the model use language?}
\begin{table}[t]
\caption{Language controls. Gap is the relative trajectory-error increase under
shuffled rather than correct text. Values are means over three seeds.}
\label{tab:text}
\centering\small
\begin{tabular}{lccc}
\toprule
Text source & Correct--shuffled gap$\uparrow$ & Seed range & Verb recall\\
\midrule
Source annotation & 2.8\% & [1.5, 4.9] & 92.9\%\\
Rule-corrected annotation & \textbf{3.1\%} & [2.3, 4.2] & 92.9\%\\
Vision-language caption & 2.9\% & [1.2, 4.0] & 47.7\%\\
Kinematic pseudo-text & 0.3\% & -- & --\\
Random label control & 0.2\% & -- & --\\
\bottomrule
\end{tabular}
\end{table}

The positive multi-seed gap in Table~\ref{tab:text}, together with near-zero gaps for
kinematic pseudo-text and random labels, suggests that FiLM captures a real semantic
signal. The effect is nevertheless small and below our 10\% target. We therefore
describe AcrossVAM1.0 as \emph{text-assisted}, not instruction-controlled. The lower
verb recall of automatically generated captions also explains why a larger language
encoder does not automatically produce a larger causal effect.

\paragraph{Cross-robot behavior.}
The language-conditioned particle model improves on Franka (+21.1\%, 136 windows),
WidowX (+19.0\%, 68), and UR5 (+18.4\%, 8), but degrades on Mobile ALOHA
($-16.1\%$, 60), Google Robot ($-5.2\%$, 32), and several small subsets. The
non-Franka aggregate remains positive (+10.0\%, 232 windows), but strict
leave-one-robot-out Mobile ALOHA experiments do not beat persistence.

\subsection{Ablations and interpretation}
\begin{table}[t]
\caption{Component ablations. Top: held-out particle trajectory-error reduction
over persistence. Bottom: delivery alternatives on the same 92-clip/460-frame
future-video test as Table~\ref{tab:video}.}
\label{tab:ablation}
\centering\scriptsize
\begin{tabular}{lc}
\toprule
Configuration & Improvement$\uparrow$\\
\midrule
Residualized boxes + appearance in dynamics & 3.9\%\\
Clean boxes, stride 1 & 4.2\%\\
Clean boxes, stride 3 & 15.1\%\\
Clean boxes, no appearance shortcut & 18.5\%\\
Final width-64, 2-layer model & \textbf{21.0\%}\\
\bottomrule
\end{tabular}
\vspace{4pt}

\begin{tabular}{lcccc}
\toprule
Delivery configuration & PSNR$\uparrow$ & SSIM$\uparrow$ & LPIPS$\downarrow$ & Motion PSNR$\uparrow$\\
\midrule
No delivery (raw generation) & 19.935 & 0.7634 & 0.1381 & \textbf{13.227}\\
Fixed particle-alpha mask & 20.168 & 0.7850 & 0.1338 & 12.931\\
Learned causal mask & \textbf{20.573} & \textbf{0.8004} & \textbf{0.1304} & 12.870\\
Pixel-selection oracle & 22.121 & 0.8327 & 0.1140 & 14.092\\
\bottomrule
\end{tabular}
\end{table}

Table~\ref{tab:ablation} summarizes the component and delivery ablations.
Raw geometric state is preferable to codec box residuals: the latter exhibit a
34.4\% velocity sign-flip rate and obscure physical displacement. Stride three
makes meaningful motion visible at 128-pixel resolution. Appearance is useful for
rendering but harmful as a dynamics shortcut. On the delivery side, fixed particle
alpha improves PSNR by 0.23 dB over raw generation, whereas the learned causal mask
adds another 0.41 dB and restores SSIM above persistence. This improvement trades
0.36 dB of motion PSNR for substantially better static-region fidelity; the oracle
shows that mask calibration still leaves 1.55 dB of PSNR headroom. Table~\ref{tab:text}
provides the complementary language ablation. Slot matching has negligible effect
in the fixed-ID cache, as expected, but remains necessary for unordered proposals.
\section{Limitations}
\label{sec:limitations}
AcrossVAM1.0 is not yet a general-purpose action model. It predicts only five future
frames at $128^2$ resolution and uses part masks during codec construction. The
frozen OpenCLIP tower is large even though the trainable dynamics are small, so
\emph{lightweight} refers to the learned predictive core rather than total deployment
memory. Global gains over persistence are modest, public-model baselines are not yet retrained on the identical VRS split, LPIPS is worse, language
sensitivity is limited, and cross-embodiment generalization is inconsistent.
The pixel oracle also reveals delivery calibration as a major bottleneck. Future
work should learn uncertainty-aware blending, collect counterfactual instruction
pairs, replace mask supervision with proposal-time segmentation, and evaluate
closed-loop control rather than video metrics alone.

\section{Conclusion}
We introduced AcrossVAM1.0, a compact text-assisted world model that treats robot video
prediction as semantic particle dynamics plus causal appearance synthesis. The
model improves trajectories and motion-region prediction while residual delivery
recovers much of the static visual fidelity. Strong controls show both what
works---low-dimensional motion modeling---and what remains unresolved---perceptual
delivery, language dependence, and cross-robot transfer. This separation provides
a foundation for interpretable video action models without hiding failure modes
behind aggregate image metrics.

\FloatBarrier
\bibliography{references,references_2026}

@inproceedings{babaeizadeh2021fitvid,
  title={FitVid: Overfitting in Pixel-Level Video Prediction},
  author={Babaeizadeh, Mohammad and Saffar, Mohammad and Nair, Suraj and Levine, Sergey and Finn, Chelsea and Erhan, Dumitru},
  booktitle={International Conference on Learning Representations}, year={2022}}

@inproceedings{carion2020detr,
  title={End-to-End Object Detection with Transformers},
  author={Carion, Nicolas and Massa, Francisco and Synnaeve, Gabriel and Usunier, Nicolas and Kirillov, Alexander and Zagoruyko, Sergey},
  booktitle={European Conference on Computer Vision}, year={2020}}

@inproceedings{chen2024diffusionforcing,
  title={Diffusion Forcing: Next-token Prediction Meets Full-Sequence Diffusion},
  author={Chen, Boyuan and Monso, Diego Marti and Du, Yilun and Simchowitz, Max and Tedrake, Russ and Sitzmann, Vincent},
  booktitle={Advances in Neural Information Processing Systems}, year={2024}}

@inproceedings{cherti2023openclip,
  title={Reproducible Scaling Laws for Contrastive Language-Image Learning},
  author={Cherti, Mehdi and Beaumont, Romain and Wightman, Ross and Wortsman, Mitchell and Ilharco, Gabriel and Gordon, Cade and Schuhmann, Christoph and Schmidt, Ludwig and Jitsev, Jenia},
  booktitle={IEEE/CVF Conference on Computer Vision and Pattern Recognition}, year={2023}}

@article{daniel2023ddlp,
  title={Deep Dynamic Latent Particles for Unsupervised Object-Centric Video Prediction},
  author={Daniel, Tal and Tamar, Aviv},
  journal={Transactions on Machine Learning Research}, year={2024}}

@inproceedings{du2023unipi,
  title={Learning Universal Policies via Text-Guided Video Generation},
  author={Du, Yilun and Yang, Sherry and Dai, Bo and Dai, Hanjun and Nachum, Ofir and Tenenbaum, Joshua B. and Schuurmans, Dale and Abbeel, Pieter},
  booktitle={Advances in Neural Information Processing Systems}, year={2023}}

@article{ebert2018visual,
  title={Visual Foresight: Model-Based Deep Reinforcement Learning for Vision-Based Robotic Control},
  author={Ebert, Frederik and Finn, Chelsea and Dasari, Sudeep and Xie, Annie and Lee, Alex and Levine, Sergey},
  journal={arXiv preprint arXiv:1812.00568}, year={2018}}

@article{elsayed2022saviplus,
  title={SAVi++: Towards End-to-End Object-Centric Learning from Real-World Videos},
  author={Elsayed, Gamaleldin F. and Mahendran, Aravindh and van Steenkiste, Sjoerd and Greff, Klaus and Kipf, Thomas and Mozer, Michael C.},
  journal={arXiv preprint arXiv:2206.07764}, year={2022}}

@inproceedings{finn2016unsupervised,
  title={Unsupervised Learning for Physical Interaction through Video Prediction},
  author={Finn, Chelsea and Goodfellow, Ian and Levine, Sergey},
  booktitle={Advances in Neural Information Processing Systems}, year={2016}}

@article{hafner2023dreamerv3,
  title={Mastering Diverse Domains through World Models},
  author={Hafner, Danijar and Pasukonis, Jurgis and Ba, Jimmy and Lillicrap, Timothy},
  journal={arXiv preprint arXiv:2301.04104}, year={2023}}

@inproceedings{ho2020ddpm,
  title={Denoising Diffusion Probabilistic Models},
  author={Ho, Jonathan and Jain, Ajay and Abbeel, Pieter},
  booktitle={Advances in Neural Information Processing Systems}, year={2020}}

@article{huang2026nano,
  title={Nano World Models: A Minimalist Approach to Future Video Prediction},
  author={Huang, Siqiao and Kaushik, Partha and Chen, Michael and Pan, Hengkai and Geng, Kaiwen and Chehab, Omar and Moreno-Pino, Fernando and Simchowitz, Max},
  journal={arXiv preprint arXiv:2605.23993}, year={2026}}

@inproceedings{khazatsky2024droid,
  title={DROID: A Large-Scale In-the-Wild Robot Manipulation Dataset},
  author={Khazatsky, Alexander and Pertsch, Karl and Nair, Suraj and Balakrishna, Ashwin and Dasari, Sudeep and Karamcheti, Siddharth and Nasiriany, Soroush and Srirama, Mohan and Chen, Lawrence and others},
  booktitle={Robotics: Science and Systems}, year={2024}}

@inproceedings{kirillov2023sam,
  title={Segment Anything},
  author={Kirillov, Alexander and Mintun, Eric and Ravi, Nikhila and Mao, Hanzi and Rolland, Chloe and Gustafson, Laura and Xiao, Tete and Whitehead, Spencer and Berg, Alexander C. and Lo, Wan-Yen and Dollar, Piotr and Girshick, Ross},
  booktitle={IEEE/CVF International Conference on Computer Vision}, year={2023}}

@inproceedings{locatello2020slot,
  title={Object-Centric Learning with Slot Attention},
  author={Locatello, Francesco and Weissenborn, Dirk and Unterthiner, Thomas and Mahendran, Aravindh and Heigold, Georg and Uszkoreit, Jakob and Dosovitskiy, Alexey and Kipf, Thomas},
  booktitle={Advances in Neural Information Processing Systems}, year={2020}}

@inproceedings{perez2018film,
  title={FiLM: Visual Reasoning with a General Conditioning Layer},
  author={Perez, Ethan and Strub, Florian and de Vries, Harm and Dumoulin, Vincent and Courville, Aaron},
  booktitle={AAAI Conference on Artificial Intelligence}, year={2018}}

@inproceedings{radford2021clip,
  title={Learning Transferable Visual Models From Natural Language Supervision},
  author={Radford, Alec and Kim, Jong Wook and Hallacy, Chris and Ramesh, Aditya and Goh, Gabriel and Agarwal, Sandhini and Sastry, Girish and Askell, Amanda and Mishkin, Pamela and Clark, Jack and Krueger, Gretchen and Sutskever, Ilya},
  booktitle={International Conference on Machine Learning}, year={2021}}

@article{ravi2024sam2,
  title={SAM 2: Segment Anything in Images and Videos},
  author={Ravi, Nikhila and Gabeur, Valentin and Hu, Yuan-Ting and Hu, Ronghang and Ryali, Chaitanya and Ma, Tengyu and Khedr, Haitham and others},
  journal={arXiv preprint arXiv:2408.00714}, year={2024}}

@inproceedings{vaswani2017attention,
  title={Attention Is All You Need},
  author={Vaswani, Ashish and Shazeer, Noam and Parmar, Niki and Uszkoreit, Jakob and Jones, Llion and Gomez, Aidan N. and Kaiser, Lukasz and Polosukhin, Illia},
  booktitle={Advances in Neural Information Processing Systems}, year={2017}}

@article{wang2004ssim,
  title={Image Quality Assessment: From Error Visibility to Structural Similarity},
  author={Wang, Zhou and Bovik, Alan C. and Sheikh, Hamid R. and Simoncelli, Eero P.},
  journal={IEEE Transactions on Image Processing}, volume={13}, number={4},
  pages={600--612}, year={2004}}

@inproceedings{song2025ock,
  title={OCK: Unsupervised Dynamic Video Prediction with Object-Centric Kinematics},
  author={Song, Yeon-Ji and Kim, Jaein and Choi, Suhyung and Kim, Jin-Hwa and Zhang, Byoung-Tak},
  booktitle={Proceedings of the IEEE/CVF International Conference on Computer Vision},
  pages={11359--11368},
  year={2025}}

@inproceedings{wu2023slotformer,
  title={SlotFormer: Unsupervised Visual Dynamics Simulation with Object-Centric Models},
  author={Wu, Ziyi and Dvornik, Nikita and Greff, Klaus and Kipf, Thomas and Garg, Animesh},
  booktitle={International Conference on Learning Representations}, year={2023}}

@inproceedings{zhang2018lpips,
  title={The Unreasonable Effectiveness of Deep Features as a Perceptual Metric},
  author={Zhang, Richard and Isola, Phillip and Efros, Alexei A. and Shechtman, Eli and Wang, Oliver},
  booktitle={IEEE/CVF Conference on Computer Vision and Pattern Recognition}, year={2018}}

@inproceedings{daniel2026latent,
  title={Latent Particle World Models: Self-supervised Object-centric Stochastic Dynamics Modeling},
  author={Daniel, Tal and Qi, Carl and Haramati, Dan and Zadeh, Amir and Li, Chuan and Tamar, Aviv and Pathak, Deepak and Held, David},
  booktitle={International Conference on Learning Representations},
  year={2026},
  url={https://openreview.net/forum?id=lTaPtGiUUc}
}

@article{villar2026textocvp,
  title={{TextOCVP}: Object-Centric Video Prediction with Language Guidance},
  author={Villar-Corrales, Angel and Plepi, Gjergj and Behnke, Sven},
  journal={Transactions on Machine Learning Research},
  year={2026}
}
\bibliographystyle{iclr2027_conference}

\appendix
\section{Reproducibility Details}
\label{app:details}
\paragraph{Fixed protocol.}
All model selection uses 2,526 training and 106 validation clips. The 105
test-manifest clips are never used for checkpoint or threshold selection. Stages
contain fewer clips when they require complete RGB, mask, particle-cache, and text
records; each main table reports the effective sample count.

\paragraph{Optimization stages.}
We (i) train and freeze the particle codec; (ii) cache particles; (iii) train
transition and rollout dynamics; (iv) train the dense prior and residual refiner
using rollout particles; and (v) train the inference-safe blend mask. The dynamics
configuration has dimension 64, two layers, four heads, nine time steps, four
context steps, and three slots. Future prediction is autoregressive.

\paragraph{Leakage audit.}
The appearance encoder is evaluated once on $\vx_C$. Future RGB, masks, oracle
particles, and detail codes are removed from inference inputs and used only as
targets or upper bounds. Text shuffling occurs within the evaluation split.

\section{Additional Per-Robot Results}
\begin{table}[h]
\caption{Per-robot trajectory diagnostics. Positive is better than persistence.}
\centering\small
\begin{tabular}{lrr}
\toprule
Robot & Windows & Improvement\\
\midrule
Franka & 136 & +21.1\%\\
WidowX & 68 & +19.0\%\\
Mobile ALOHA & 60 & $-16.1\%$\\
Google Robot & 32 & $-5.2\%$\\
Fanuc Mate & 28 & $-9.0\%$\\
Sawyer & 20 & $-12.9\%$\\
UR5 & 8 & +18.4\%\\
\midrule
All non-Franka & 232 & +10.0\%\\
\bottomrule
\end{tabular}
\end{table}

\section{Broader Impact}
Video world models can support data-efficient robot learning and safer offline
analysis, but inaccurate rollouts can encourage unsafe actions if treated as
physical guarantees. AcrossVAM1.0 is a research prototype and should not be used as a
safety controller. Its data may encode embodiment, laboratory, and annotation
biases. Part-level visualization can expose failures, but does not eliminate the
need for real-world validation and appropriate safety constraints.

\end{document}